\documentclass[10pt,twocolumn,letterpaper]{article}

\usepackage{iccv}
\usepackage{times}
\usepackage{epsfig}
\usepackage{graphicx}
\usepackage{amsmath}
\usepackage{amssymb}
\usepackage{bbm}
\usepackage{booktabs}
\usepackage{tabularx}
\usepackage{booktabs}
\usepackage{multirow}
\usepackage{color, colortbl}
\definecolor{LightCyan}{rgb}{0.88,1,1}
\usepackage{makecell}%导入宏包
\usepackage{graphicx}
\usepackage{subfigure}
\usepackage{amsmath}
\usepackage{amssymb}
\usepackage{comment}
\usepackage{color}
\usepackage{url}
\usepackage{hhline}  % 双线表
\usepackage{amsmath, amssymb}
\usepackage[ruled,vlined]{algorithm2e}
\usepackage{algorithmic} %format of the algorithm 
\usepackage{multirow} %multirow for format of table 
\usepackage{amsmath} 
\usepackage{xcolor}
\usepackage{multirow}
\usepackage{float}
\usepackage{graphicx}
\usepackage[switch]{lineno}
\usepackage[pagebackref=true,breaklinks=true,letterpaper=true,colorlinks,bookmarks=false]{hyperref}

\iccvfinalcopy % *** Uncomment this line for the final submission

\def\iccvPaperID{1136} % *** Enter the ICCV Paper ID here

\ificcvfinal\fi

\begin{document}

%%%%%%%%% TITLE
\title{Contrastive and Selective Hidden Embeddings for Medical Image Segmentation}

\author{Zhuowei Li $^*$\\
	Rutgers University\\
	{\tt\small zl502@cs.rutgers.edu}
\and
Zihao Liu $^*$\\
Peking University\\
{\tt\small lzh19961031@pku.edu.cn}
\and
Zhiqiang Hu\\
SenseTime Research\\
{\tt\small huzhiqiang@sensetime.com}
\and
Qing Xia\\
SenseTime Research\\
{\tt\small xiaqing@sensetime.com}
\and
Ruiqin Xiong\\
Peking University\\
{\tt\small rqxiong@pku.edu.cn}
\and
Shaoting Zhang\\
SenseTime Research\\
{\tt\small zhangshaoting@sensetime.com}
\and
Dimitris Metaxas\\
Rutgers\\
{\tt\small dnm@cs.rutgers.edu}
\and
Tingting Jiang\\
Peking University\\
{\tt\small ttjiang@pku.edu.cn}
}

\maketitle
\ificcvfinal\thispagestyle{empty}\fi

\def\thefootnote{*}\footnotetext{Equal contributions}

%%%%%%%%% ABSTRAC
\begin{abstract}
Medical image segmentation has been widely recognized as a pivot procedure for clinical diagnosis, analysis, and treatment planning. However, the laborious and expensive annotation process lags down the speed of further advances. Contrastive learning-based weight pre-training provides an alternative by leveraging unlabeled data to learn a good representation. In this paper, we investigate how the contrastive learning benefits the general supervised medical segmentation tasks. To this end, patch-dragsaw contrastive regularization (PDCR) is proposed to perform patch-level tugging and repulsing with the extent controlled by a continuous affinity score. And a new structure dubbed uncertainty-aware feature selection block (UAFS) is designed to perform the feature selection process, which can handle the learning target shift caused by minority features with high-uncertainty. By plugging the proposed 2 modules into the existing segmentation architecture, we achieve state-of-the-art results across 8 public datasets from 6 domains. Newly designed modules further decrease the amount of training data to a quarter while achieving comparable, if not better, performances. From this perspective, we take the opposite direction of the original self/un-supervised contrastive learning by further excavating information contained within the label. Codes are publicly available at \url{https://github.com/lzh19961031/PDCR_UAFR-MIS}.

\end{abstract}

\section{Introduction}
Medical image segmentation has been widely recognized as a pivot procedure for clinical diagnosis, analysis, and treatment planning. Despite the breakthroughs introduced by deep learning techniques in recent years, the ethics-respected data collection process and expertise-demanding annotation procedure makes it inefficient to improve the segmentation quality by simply adding more data. Further exploring unlabeled data (un-/self-supervise learning) and excavate underlying information from labeled data (supervised learning) are two orthogonal methodologies to plumb.

\begin{figure}[t]
    \centering
    \includegraphics[width=0.8\linewidth]{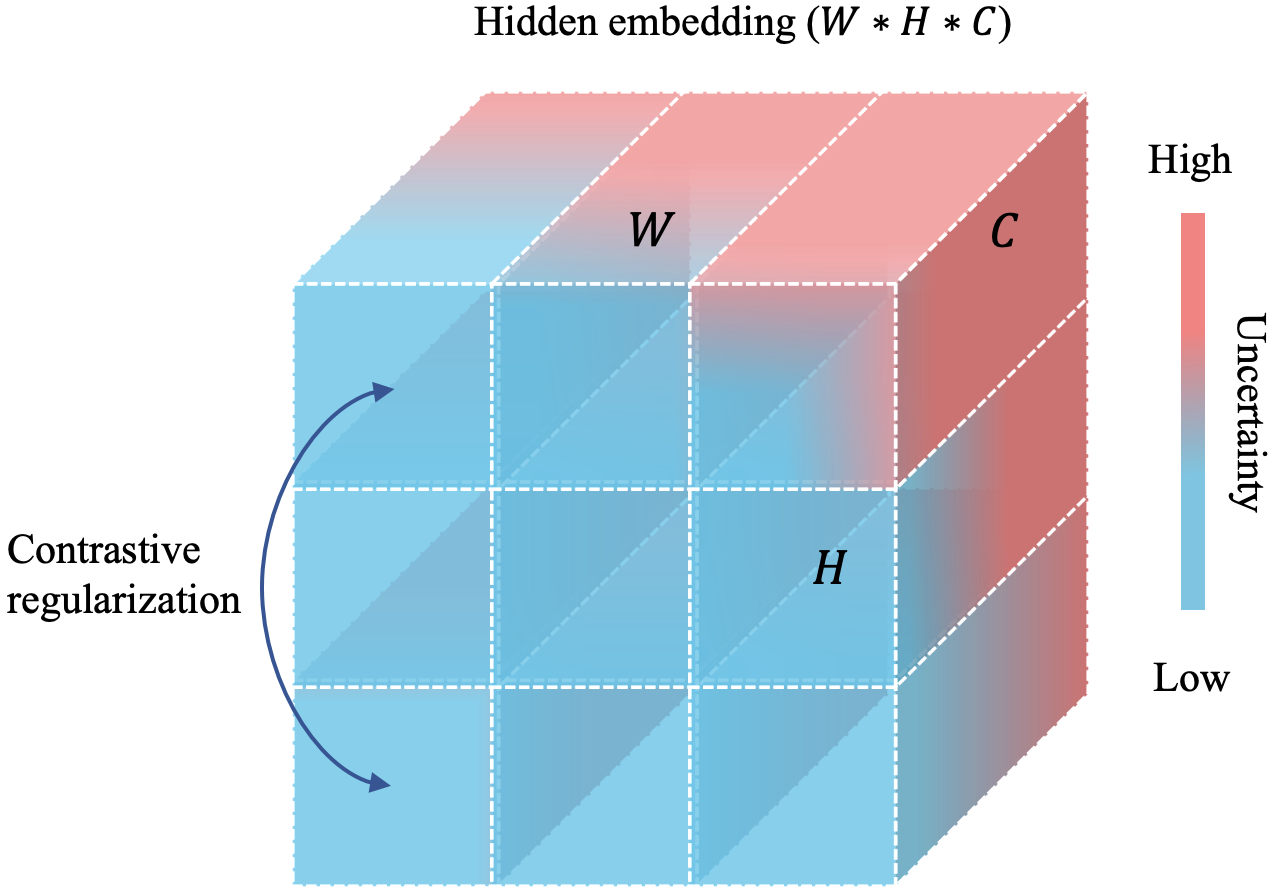}
    \caption{In a hidden feature map, a continuous constrastive regularization is imposed among embedded vectors and the uncertainty is measured as a proxy for vector-wise dynamic selection.}
    \label{img:concept}
\end{figure}

Most recently, contrastive learning has been proved to have great potentials in un-/self-supervised representation~\cite{CL3,CL4,CL4.5,CL5,CL6,CL7,CL8,CL10,CL12} by pulling together different views (augmentations) of the same image (positive pairs) while pushing away distinct images (negative pair). Beyond this unsupervised scenario, contrastive regularization also demonstrates its promotional effect in supervised classification~\cite{sup_cl} through extending the definition of ”negative pair” from inter-image to inter-category. Although efforts have also been made to fit the contrastive learning to the segmentation framework~\cite{cl_seg1,cl_seg2,cl_seg3}, existing works are scoped under the self-supervised weight pre-training regime which mainly explores the inter-image mutual information maximization. In this work, we investigate how the contrastive regularization can benefit general medical image segmentation tasks in the supervised fashion. To approach this, we first redefine the fundamental concept, \textbf{\textit{pair}}, as it is tailored for inter-image entities beforehand, and it does not fully respect the properties of the segmentation task which is a pixel-level classification in essence. Specifically, we observe the following 2 contradictions:

\begin{itemize}
    \item Instead of inter-image discrimination, the segmentation task focuses on intra-image discrimination.
    So we hypothesis that a pair should involve two entities within the same image instead of acrossing images. 
    \item Although the pixel labels are discriminated discretely, the distribution of pixel values and the associated features are continuous and smooth. Therefore, it is likely that a rigid and mutual-exclusive bipartition of either ``positive" or ``negative" will lead to the sub-optimum.
\end{itemize}

Based on these observations and assumptions, we propose the \textbf{\textit{patch-dragsaw contrastive regularization}} (\textit{PDCR}) that explores the patch-level contrastive constraints without the loss of the continuity of the image distribution. In our design, pairs are coupled directly from hidden feature spaces. Referencing the concept of the receptive field, this essentially model a pair as two local image patches that may overlap. Since the patches may contain both foregrounds and backgrounds, we propose to measure the \textit{foreground ratio} proximity as a pairwise similarity metric for contrastive learning. The core idea here is the two patches with similar constitutions in the original image space should converge to less divergence in feature spaces regardless of their geometries. As a result, pairs are not rigidly bipartite as either positive or negative. Instead, forces of tugging and repulsing are coexisting, with the extent controlled by the proposed similarity measure termed \textbf{\textit{affinity score}}. This can be figuratively described as a \textit{dragsaw}. The \textit{patch-dragsaw} constraint strikes for a balance that respects the continuous nature of the image distribution as well as exploiting the contrastive power.

\textit{Patch-dragsaw} exploits inter-vector relations within a feature map while treating vectors identically. However, due to ambiguous boundaries and heterogeneous textures in medical contexts, the minority of pixels/voxels with high-uncertainty may suffice to mislead the network optimization, given majorities with high-confidence. As a remedy, \textbf{\textit{uncertainty-aware feature selection}} (UAFS) is designed to model the vector-wise uncertainty in the feature space and then apply it back to function as a feature selection process. Here, we leverage the network itself to model the uncertainty, and no extra supervision is needed. 

In a nutshell, two modules are proposed in this work to explore inter-vector relatives and perform vector-wise dynamic selection, respectively. This process can be simplified as in Fig~\ref{img:concept}. It is worth mentioning that both modules are designed and implemented in a ``lightweight plugin" fashion, meaning that they can be easily integrated into any existing architectures, segmentation frameworks, or training pipelines. We show that a task-agnostic design, with the innovations plugged into DeepLabv3+~\cite{DeeplabV3+} for 2D segmentation and nnUNet~\cite{nnUNet} for 3D segmentation, can achieve consistent advantages over other task-tailored counterparts, across 8 public datasets from 6 domains. The empirical results also demonstrate that the two techniques are complementary to each other. Our main contributions can be summarized in threefold:
\begin{itemize}
    \item \textit{Patch-dragsaw contrastive regularization} is proposed to extend the contrastive learning to the supervised segmentation task by redefining multi-scale patch-level pairs and relaxing the definition of a discrete and mutual exclusive bipartition into a continuous and soft weight. 
    \item \textit{Uncertainty-aware feature selection} module is designed to leverage the network itself to generalize the feature selection process in order to against learning-attention shift issue caused by misleading features with high uncertainty.  
    \item State-of-the-art results have been achieved across 8 diverse datasets from 6 domains using a task-agnostic design. Beyond this blend improvement, two designs are substantiated to boost the semi-supervised learning scenario greatly. Our method outperforms the baseline model trained with full data using only a quarter.   
\end{itemize}
%-------------------------------------------------------------------------------------------------------------------------
\section{Related Work}

\subsection{Architectures for Medical Image Segmentation}
A vast of neural networks have been designed and tuned for medical image segmentation. Despite the unique design of each architecture, most of the networks employ an encoder-decoder structure with the bypass connection to fuse low-level features with high-level features. Among those designs, FCN~\cite{FCN}, UNet~\cite{UNet} and DeepLab~\cite{DeeplabV3+} function as three milestones. They and their variations provide a stable and consistent baseline for segmentation tasks.

In recent years, nnUNet~\cite{nnUNet} shows that network modification may be inferior compared with searching suitable hyper-parameters including crop-sizes, normalization, augmentations, pre-processing steps, post-processing techniques, etc. Its pipeline and code-base provide a strong baseline on a broad scope of tasks. In this work, benefits gained by adding proposed techniques to 3 milestone architecture as well as nnUNnet pipeline can be viewed as a sign indicating that there is still room to improve in network design and information digging. 

\subsection{Contrastive Learning}
Contrastive learning-based model pretrain has largely bridged the gap between supervised and unsupervised model pretraining~\cite{CL3,CL4,CL4.5,CL5,CL7,CL8} by learning to discriminate positive pair against negative pairs. SimCLR~\cite{CL5} demonstrates the importance of augmentations and shows it is beneficial to maintain a large number of negative pairs and introducing the projection neck. Moco~\cite{CL4,CL4.5} on the other hand, utilizes the memory bank to eliminate the limitation caused by batch-size. BYOL~\cite{CL6} and SimSiam~\cite{CL12} further abandon negative pairs and exploring the role of the siamese network in representation learning.

Besides backbone pretraining, efforts have also been made to fit contrastive learning into segmentation tasks. \cite{cl_seg2} leverages global contrastive loss and local contrastive loss to pretrain encoder and decoder, respectively. \cite{cl_seg1} proposes a pixel-wise contrastive loss for segmentation pretraining and exploring advantages gained under the semi-supervised scenario. Patch-dragsaw loss proposed in this task is the first technique, to our best knowledge, that boosts supervised segmentation by contrastive learning.    
\begin{figure*}
    \centering
    \includegraphics[width=0.85\textwidth]{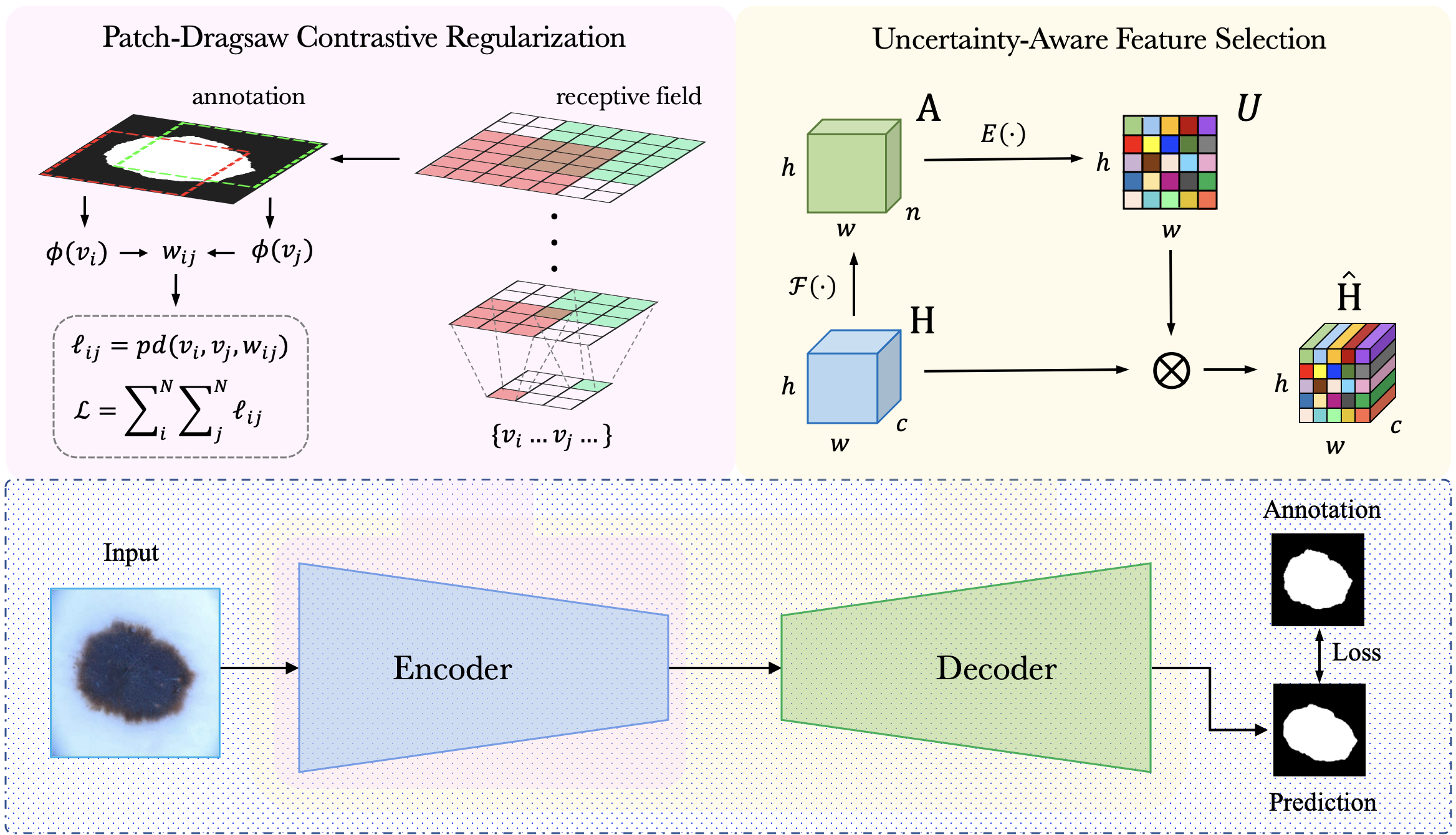}
    \caption{An overview of proposed methods. PDCR is deployed at the encoder and UAFS can be inserted at both encoder and decoder. Both blocks in upper region of the figure are illustrations of designed methods for a single layer. For PDCR, $\phi(\cdot)$ calculate foreground ratio for a vector according to its corresponding patch. $w_{ij}$ measures similarity between vector $v_i$ and $v_j$. $pd(\cdot)$ calculate patch-dragsaw loss for a pair. For UAFS, $\mathcal{F}(\cdot)$ perform segmentation at specific scale and $E(\cdot)$ represents softmax function following by shannon-entropy across channels.}
    \label{framework}
\end{figure*} 

\subsection{Uncertainty-guided segmentation}
Several uncertainty-based methodologies are introduced for image segmentation~\cite{Uncertainty1,Uncertainty5,Uncertainty3}. 
According to~\cite{Uncertainty1}, uncertainty can be divided into two branches in deep learning, aleatoric uncertainty, which is caused by data itself and often occurred in boundaries of objects, and epistemic uncertainty, which is attributed to limited data and insufficient model training. Both uncertainties, regardless of sources, are then utilized to boost the segmentation process. 
~\cite{Uncertainty5} quantifies epistemic uncertainty by running a given sample through the model several times with the dropout layers.
Here we leverage entropy to measure the uncertainty that can be classified as either aleatoric uncertainty or epistemic uncertainty. Features with high uncertainty are then deemed as detrimental and thus be suppressed.

%-------------------------------------------------------------------------------------------------------------------------
\section{Patch-Dragsaw Contrastive Regularization}
In this section, we first give a brief introduction to self-supervised contrastive learning. Then we detailed two fundamental concepts of our innovation. Finally, we come up with the \textit{patch-dragsaw contrastive regularization}. The Upper left of Fig.~\ref{framework} provides a straightforward view of the \textit{PDCR}.

\subsection{Self-Supervised Contrastive learning}\label{SSCL}
Self-supervised contrastive learning framework generally contains three principle components: (i) a transformation collection $\mathcal{T}$ which commonly contains \textit{random cropping with resizing back}, \textit{color distortion}, \textit{Gaussian blur} and etc. (ii) an encoder network $E(\cdot)$ to encode high-dimensional inputs. (iii) a projection head $P(\cdot)$ to further introduce non-linearity and reduce the dimension of output embeddings. 

As in~\cite{CL5}, a positive pair $(\tilde{\boldsymbol{x}}_{i}, \tilde{\boldsymbol{x}}_{j(i)})$ is acquired by applying randomly sampled augmentations $t \sim \mathcal{T}$ of the same image $x$ twice. Since $(\tilde{\boldsymbol{x}}_{i}, \tilde{\boldsymbol{x}}_{j(i)})$ and $(\tilde{\boldsymbol{x}}_{j(i)}, \tilde{\boldsymbol{x}}_{i})$ are treated as two pairs, a minibatch with size $N$ will attribute to $2N$ pairs for training. Then the objective function managing to minimize is defined as:
\begin{equation}
    \mathcal{L}_{batch} = \sum^{2N}_{i=1}\mathcal{L}_{i}
\end{equation}
\begin{equation}
    \mathcal{L}_i = -\log{\frac{\exp{(sim(\boldsymbol{z}_i,\boldsymbol{z}_{j(i))}/\tau)}}{\sum^{2N}_{k=1}\mathbbm{1}_{[k \neq i]}\exp(sim(\boldsymbol{z}_i,\boldsymbol{z}_k))/\tau)}}
\end{equation}
where $\boldsymbol{z}_i = P(E(\tilde{\boldsymbol{x}}_{i}))$, $\mathbbm{1}_{[k \neq i]}\in\{0,1\}$ returns 1 iff $k \neq i$ and $sim(\cdot)$ denotes the cosine similarity between two vectors.

\subsection{Patch-Pair for Segmentation}
\label{PP}
To explore intra-image relatives, we construct patch-level pairs within an image rather than employing inter-image instances. Instead of consuming cropped patches explicitly, we leverage the \textit{receptive field} concept to correlate local patches from the input space with hidden vectors in the hidden feature maps. So the proposed framework only comprises a single fundamental component, an encoder. To improve the training efficiency and fit the modern GPU memory, we perform \textit{grid-sampling} to select $N$ hidden vectors, denoted as $\boldsymbol{v}$, from each hidden layer. Note that we do not fork the entity by random augmentations, and we assume all entities as both \textit{positive} and \textit{negative} for each other. So by sampling $N$ vectors within a layer will result in $N^2$ densely connected pairs in total. Since adding proposed regularization at different layers attributes to different patch-sizes, we can easily change the patch-size or combine multi-scale patches by manipulating locations where the regularization is applied.

\subsection{Affinity Score}
\label{AS}
Given a pair of vectors $(\boldsymbol{v}^{i}, \boldsymbol{v}^{j})$, the associated affinity score $w_{ij} \in [0,1]$ is a scalar that measures to what extent should a pair be \textit{pulling} and \textit{pushing}. 
Denote the receptive field of $\boldsymbol{v}^{i}$ as $R_i$.
Formally, $w_{ij}$ is calculated as:

\begin{equation}
w_{ij}=1-\frac{1}{M}\sum^{M}_{m=1}|\phi^i_m - \phi^j_m|
\label{eq:as}
\end{equation}
where $M$ denotes the number of classes and $\phi^i_m$ is the foreground ratio of class $m$ for vector $\boldsymbol{v}^{i}$. 
$\phi^i_m$ measures the area ratio of class $m$ compared with total area in the corresponding receptive field and it takes the following form:

\begin{equation}
\phi^i_m = \frac{ \text{Number of pixels in}  \ R_i  \ \text{which belongs to class } m}  { \text{Area of receptive field} \  R_i} 
\label{eq:fr}
\end{equation}

Specifically, the \textit{foreground-ratio} $\phi_m^i$ is computed as following. Suppose $\boldsymbol{v}^{i}$ is sampled from layer $L$. The size of receptive-field $R_i$ is denoted as $r_{L}$ , which is calculated according to:
 \begin{equation}
   r_{L}=\sum_{l=1}^{L}\left((k_l - 1)\prod_{t = 1}^{l - 1}s_t\right)+1 \label{eq:rf}
 \end{equation}
 
where $k$ and $s$ denote the kernel size and the stride size of a convolutional layer, respectively. Note that the size of receptive field $r_{L}$ only depends on layer $L$, not related to specific vectors. The area of $R_i$ is equal to ${r_{L}}^2$, which is the denominator in Eqn.~\ref{eq:fr}.
Given the locations of vector $\boldsymbol{v}^{i}$ in the hidden feature, we can calculate the location of $R_i$ in the original image according to~\cite{rf}.
With the calculated location of $R_i$, the number of pixels which belongs to each class $m$ in $R_i$ can be computed according to the image annotation $G$, which is the numerator in Eqn.~\ref{eq:fr}.

\subsection{Patch-Dragsaw Contrastive Loss}
Referencing the statement of \textit{patch-pair} Sec~\ref{PP} and \textit{affinity score} Sec~\ref{AS}, the proposed \textbf{\textit{Patch-dragsaw}} contrastive loss takes the following formation:
\begin{equation}
s_{ij} = sim(\boldsymbol{v}_i, \boldsymbol{v}_j) = \frac{\boldsymbol{v}_i^T \boldsymbol{v}_j}{\left \|{\boldsymbol{v}_i} \right \|  \left \|{\boldsymbol{v}_j} \right \|}
\end{equation}
\begin{equation}
    \mathcal{L}_l = \sum^{N}_{i=1}\sum^{N}_{j=1}-\log\frac{exp([s_{ij}\cdot w_{ij}]/\tau)}{\sum^{N}_{k=1}exp([s_{ik}\cdot \overline{w}_{ik}]/\tau)}
\end{equation} 
where $\boldsymbol{v}$ is the hidden vector from the target layer $l$, $w_{ij}$ measures similarity, and  $\overline{w}_{ij} = 1 - w_{ij}$ which measures dissimilarity .
 The proposed loss essentially encourage the network to attract local-patches with similar semantics regardless of their physical locations while distinguishing patches that sharing no interest contents. Algorithm~\ref{algorithm} provides an overview of the proposed method. 

\begin{algorithm}
	\SetAlgoLined
	\SetKwInOut{Input}{input}
	\Input{encoder $E(\cdot)$, constant $N$ and $\tau$}
	%\KwResult{Network Parameters $\theta_{ag}$, $\theta_{sl}$ and $\theta_c$ }
	\For{image $I$, annotation $G$ in dataset} {
		\For{layer $l \in \{1,...,L\}$ in encoder}{
			sampling $\{\boldsymbol{v}_i,...,\boldsymbol{v}_N\}$ \\
			calculate receptive field according to Eq.~\ref{eq:rf} \\ 
			\For{i  $ \in \{1, ... , N\}$}{
							\For{j  $ \in \{1, ... , N\}$}{compute $\phi$  referencing Eq.~\ref{eq:fr}\\
								compute $w_{ij}$ with Eq.~\ref{eq:as}  \\
								$\overline w_{ij}$ =  1 - $w_{ij}$ \\
								$s_{ij} = \frac{\boldsymbol{v}_i^T \boldsymbol{v}_j}{\left \|{\boldsymbol{v}_i} \right \|  \left \|{\boldsymbol{v}_j} \right \|}$\\ $\ell_{ij}=-\log\frac{exp([s_{ij}\cdot w_{ij}]/\tau)}{\sum^{N}_{k=1}exp([s_{ik}\cdot \overline{w}_{ik}]/\tau)}$\\
				}
			} 
			$\mathcal{L}_l$ = $\sum_{i=1}^N$ $\sum_{j=1}^N$ $\ell_{ij}$ \\
		}
		$\mathcal{L}$ = $\frac{1}{L}\sum_{l=1}^{L}\mathcal{L}_l$ \\
		update network $E$ to minimize $\mathcal{L}$
	}
	\caption{Patch-dragsaw contrastive loss.}
	\label{algorithm}
\end{algorithm}

\begin{table*}[t]
\footnotesize
\begin{center}
\begin{tabular}{cccccc}
\toprule
Dataset & Domain & Task & Dimension & Modality & Evaluation Protocol \\
\midrule
ISIC 2016~\cite{ISIC2016}+PH2~\cite{PH2} & Skin & Lesion segmentation & 2D & Dermoscopy & 900 train, 200 test (Official split) \\
ISIC 2016~\cite{ISIC2016} & Skin & Lesion segmentation & 2D & Dermoscopy & 900 train, 379 test (Official split) \\
ISIC 2017~\cite{ISIC2017} & Skin & Lesion segmentation & 2D & Dermoscopy & 2000 train, 600 test (Official split)  \\
MC~\cite{MC} & Lung & Lung segmentation & 2D & X-Ray & 80 train, 58 test \\
DigestPath~\cite{DigestPath} & Pathology & Lesion segmentation & 2D & Pathology & 952 train, 5-fold cross-validation \\
Decathlon~\cite{Decathlon}-Task01 & Brain & Tumour segmentation & 3D & MRI & 454 train, 5-fold cross-validation (Official split)\\
Decathlon~\cite{Decathlon}-Task02 & Heart & Heart segmentation & 3D & MRI & 20 train, 5-fold cross-validation (Official split)\\
Decathlon~\cite{Decathlon}-Task05 & Prostate & Prostate segmentation & 3D & MRI & 32 train, 5-fold cross-validation (Official split)\\
\bottomrule
\end{tabular}
\end{center}
\caption{Details of tasks and datasets used in experiments.}\label{datasets}
\end{table*}

\begin{table*}[t]
\footnotesize
\begin{center}
\begin{tabularx}{500pt}{l>{\centering}X>{\centering}Xl>{\centering}X>{\centering}X>{\centering}Xl>{\centering}X>{\centering}X>{\centering}Xl>{\centering}XX}
\toprule
\multicolumn{3}{c}{ISIC 2016+PH2 (Skin)} & \multicolumn{4}{c}{ISIC 2016 (Skin)} & \multicolumn{4}{c}{ISIC 2017 (Skin)} &\multicolumn{3}{c}{MC (Lung)}\\
\cmidrule(lr){1-3} \cmidrule(lr){4-7} \cmidrule(lr){8-11} \cmidrule(lr){12-14}
Methods & DI & JA  &Methods &  DI & JA & AC & Methods & DI & JA & AC &  Methods  & JA& AC \\
\midrule
MSCA~\cite{MSCA}  &  81.57 & 72.33 &DFCN~\cite{Yuan2017}  & 91.20 & 84.70 & 95.50 & FCN+SSP~\cite{FCN+SSP}  & 85.7 & 77.3& 93.8 & FCN~\cite{FCN} & 90.53& 97.35 \\
SSLS~\cite{SSLS} & 78.38 & 68.16  & MSFCN~\cite{Bi2017}  & 91.18 & 84.64& 95.51 & Bi et al.~\cite{Bi2019}  & 85.66 & 77.73& 94.08 &  UNet~\cite{UNet} & 91.64& 97.82 \\
FCN~\cite{FCN} & 89.40 & 82.15  &ECDN~\cite{Yuan20172} & 91.30 & 84.90 & 95.70 & SLSDeep~\cite{SLSDeep}  & 87.8& 78.2 & 93.6 &  M-Net~\cite{MNet} & 91.95& 97.96 \\
Bi et al.~\cite{Bi2017} & 90.66 & 83.99 & Bi et al.~\cite{Bi2019}& 91.77 & 85.92  & 95.78 & MBDCNN~\cite{MBDCNN}  & 87.8 & 80.4& 94.7  &  Multi-task~\cite{Multitask} &  92.24& 98.13  \\
SBPS~\cite{SBPS} & 91.84 & 84.30 &biDFL~\cite{biDFL} & 93.33& 88.12  & 96.75 & biDFL~\cite{biDFL}  &  88.54& \bf 81.47 & 94.65 &  ETNet~\cite{ETNet} & 94.20& 98.65   \\
\rowcolor{LightCyan}
Ours &  \bf 92.90 & \bf 85.89& Ours  & \bf 94.18 & \bf 88.27& \bf 97.16 & Ours  &\bf  88.74 &80.86& \bf 95.18 & Ours  & \bf 94.70& \bf 99.06  \\

\bottomrule
\end{tabularx}
\begin{tabularx}{500pt}{l>{\centering}X>{\centering}X>{\centering}Xl>{\centering}X>{\centering}X>{\centering}X>{\centering}X>{\centering}X>{\centering}X>{\centering}XX}
\toprule
\multicolumn{4}{c}{\multirow{2}{*}{DigestPath (Pathology)}} & \multicolumn{9}{c}{Decathlon}\\
\cmidrule(lr){5-13}
&&&& & \multicolumn{4}{c}{Brain Tumor} & \multicolumn{1}{c}{Heart} &\multicolumn{3}{c}{Prostate}\\
\cmidrule(lr){1-4} \cmidrule(lr){6-9} \cmidrule(lr){10-10} \cmidrule(lr){11-13}
Methods  & DI & JA& AC & Methods & ED & NE & EN & Avg & LA & PE & TR & Avg\\
\midrule
 FCN~\cite{FCN} & 77.98 & 64.07 &  94.32 & U-ResNet~\cite{UResNet} & 79.10 & 58.38  & 77.37 & 71.61 & 91.48 & 48.37 & 79.17 & 63.77\\
UNet~\cite{UNet} & 77.33 & 63.17& 94.29 & nnUNet NoDA~\cite{nnUNet} & 81.27 & 60.92 & 77.90 & 73.36 & 92.85  & 58.61 & 83.61 & 71.11\\
Dilated-Net~\cite{DilatedNet}& 77.34  & 63.45& 94.27& nnUNet~\cite{nnUNet}  & 81.68 & 61.29 & 77.97 & 73.65 & 93.21 & 63.14 &86.53 & 74.84\\
DeeplabV3+~\cite{DeeplabV3+}  & 77.92 & 63.88& 94.33 & SCNAS~\cite{SCNAS} & 80.41 & 59.85 & 78.50 & 72.92 & 91.91 & 53.81 &82.02 & 67.92\\
-&-&-&- & ASNG~\cite{ASNG} & 81.99 & 62.06 & \bf78.58 & 74.21 & 93.36 & 67.65 &  87.04 & 77.35\\
\rowcolor{LightCyan}
Ours & \bf 79.33  &\bf  65.36& \bf 95.75& Ours &\bf82.41 &\bf63.29 &78.23 &\bf74.64 & \bf 93.50 & \bf 67.74& \bf 88.02& \bf 77.88  \\
\bottomrule
\end{tabularx}
\end{center}
\caption{Results of the proposed model against state-of-the-art methods on a broad scope of medical image segmentation tasks.}\label{benchmark}
\end{table*}
%----------------------------------------------------------------
\section{Uncertainty-Aware Feature Selection}
\subsection{What feature to emphasis?}
Under most scenarios, the neural network will capture the principle semantics of an image by updating parameters through back-propagation technique. However, along with the training process, the network naturally shifts the learning attention from major semantics to everlasting \textit{"hard cases"} that generates high gradients. The question then emerges as: Does fitting \textit{hard} cases as others truly benefit the overall segmentation result? Our assumption here is quite counter-intuitive. Instead of performing hard-case mining strategy as what most methods are focusing on. We assume that minority of features with high-uncertainty may function adversely. This can be particularly true for the medical domain where the boundaries of organs or lesions are blurred and pixel-level annotations for the boundary region can not be fully accurate. So we want to courage the model to focus on "large-scale" semantics than the detailed pixel-level accuracy. To this end, we cover an \textit{\textbf{uncertainty-aware}} selection map on the hidden embeddings across all channels in order to highlight salient regions while suppressing gloomy parts. 

\subsection{Entropy-Guided Feature Selection}
Given a hidden feature $H \in \mathbb{R}^{H \times W \times C}$. Three steps are taken to build the proposed module. We first perform segmentation at specific scale:
\begin{equation}
    A = softmax(\mathcal{F}(H))
\end{equation}
Here, $\mathcal{F}(\cdot)$ is a $3\times3$ conv layer followed by an identity conv layer. Relu activation function and batch-normalization are used in between. $A\in \mathbb{R}^{H \times W \times M}$ is then the segmentation logits where $M$ is number of classes of the task. 

The uncertainty $u_{ij}$ for each location $\boldsymbol{a}_{ij} \in \mathbb{R}^{1 \times M}$ in $A$ is modeled according to normalized \textit{shannon-entropy}~\cite{entropy}:

\begin{equation}
u_{ij} = - \sum_{m}^M  \frac{\boldsymbol{a}_{ij}^m  log_2 (\boldsymbol{a}_{ij}^m)}{log_2(M)}. 
\end{equation}

Finally the uncertainty-aware feature selection map $\hat{U} \in \mathbb{R}^{H \times W}$ is acquired and imposed back to the hidden feature $H$ as following:

\begin{equation}
    \hat{U} = 1-U, \qquad \hat{H} = H \ast (1+\hat{U})
\end{equation}

The new hidden features $\hat{H}$ is then passed to the following layers instead of the original one $H$.
%-------------------------------------------------------------------------------------------------------------------------
\begin{figure*}[t]
    \centering
    \includegraphics[width=\textwidth]{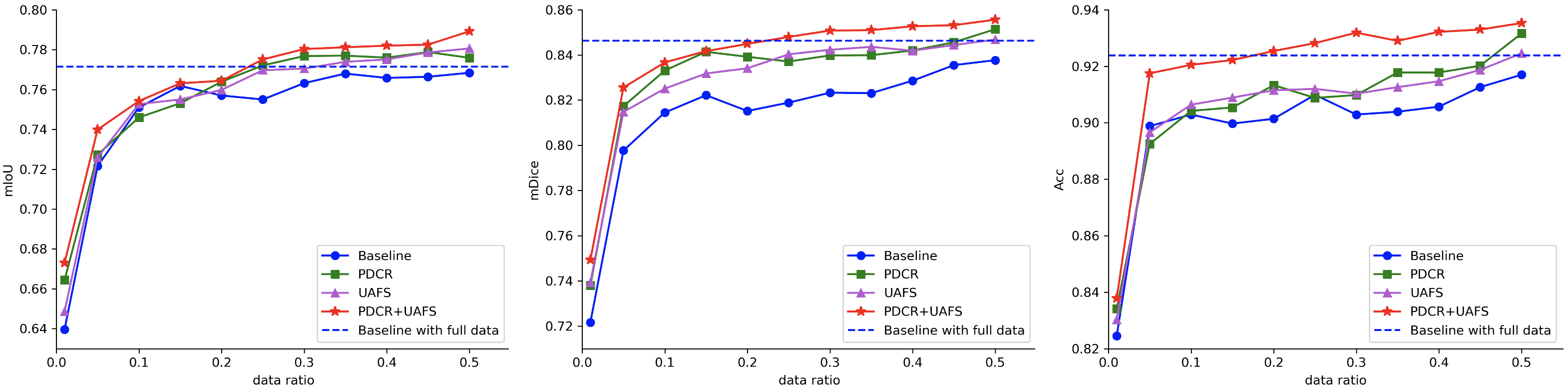}
    \caption{Comparison of proposed two techniques against the baseline with varied training data sizes. Note that model equipped with both techniques surpasses the baseline model trained with all data by using only 25\% data.}
    \label{img:semi}
\end{figure*}

\section{Experiments}
\subsection{Benchmark Studies} 
To better evaluate the generalization on a broad scope of medical image segmentation tasks, we conduct experiments on both 2D and 3D datasets,  which contain eight datasets across six domains and five modalities in total. The performance is measured by Jaccard index (JA)~\cite{JA}, Dice coefficient (DI)~\cite{DI}, and pixel-wise accuracy (AC). Table~\ref{datasets} elaborates task details\footnote{The code will be released upon publication. }. 

For benchmark studies we utilize DeepLabV3+~\cite{DeeplabV3+} (2D) or nnUNet~\cite{nnUNet} (3D) as the underlying architecture. Patch-dragsaw contrastive regularization is equipped at the last layer of block 2, 3, and 4 in the encoder, according to the ablation study. Uncertainty-aware feature selection, on the other hand, is added to all encoder and decoder layers (except the neck module in DeeLabv3+). Other hyper-parameters, including learning rate scheduler, augmentation strategy, weight decay and so on are kept the same across all 2D or 3D tasks. For 2D tasks, we utilize the Adam~\cite{adam} optimizer with the base learning rate 0.001 and cosine scheduler. The weight decay is set to 0.00001. Applied Augmentation includes scale, flip, rotate, shift, and shear transformation. For 3D tasks, the hyper-parameters follow the default settings of nnUNet. It is hypothesized that the performance still has room to improve with carefully tuned hyper-parameters for each task, respectively.

Table~\ref{benchmark} shows the results for the proposed model against other state-of-the-art methods for the aforementioned medical image segmentation tasks. It is encouraging to see that our model, with a task-agnostic design, achieves a consistent advantage over other task-tailored counterparts. We believe it is evidence that the proposed contrastive guidance and uncertainty modeling exhibit universal applicability for the medical image segmentation research area. 

\subsection{On Limited Training Data}
As mentioned before, medical segmentation models are constrained from further improvements by simply adding more data. So effective training with limited trainset sizes has been widely accepted as one of the feasible solutions for the problem. Both PDCR and UAFS are designed with this in mind. In this subsection, we explore the robustness of proposed methods against attack from data reduction on ISIC2017. Specifically, we gradually decrease the amount of training data from 50\% of the whole set to 1\% with a stride of 5\% and see the performance changes along the way. The results are shown in Fig.~\ref{img:semi}. It is satisfactory to see that the performance is less affected by the data size in a broad range from 25\% to 100\%. Noticeably, the model still performs on par with the whole-set baseline with as little as 20\% of overall training data. This property reveals the great potential of proposed methods in handling other especially long-tailed domains where purely big-data-driven methods may not be applicable.
\begin{figure}[t]
    \centering
    \includegraphics[width=0.9\linewidth]{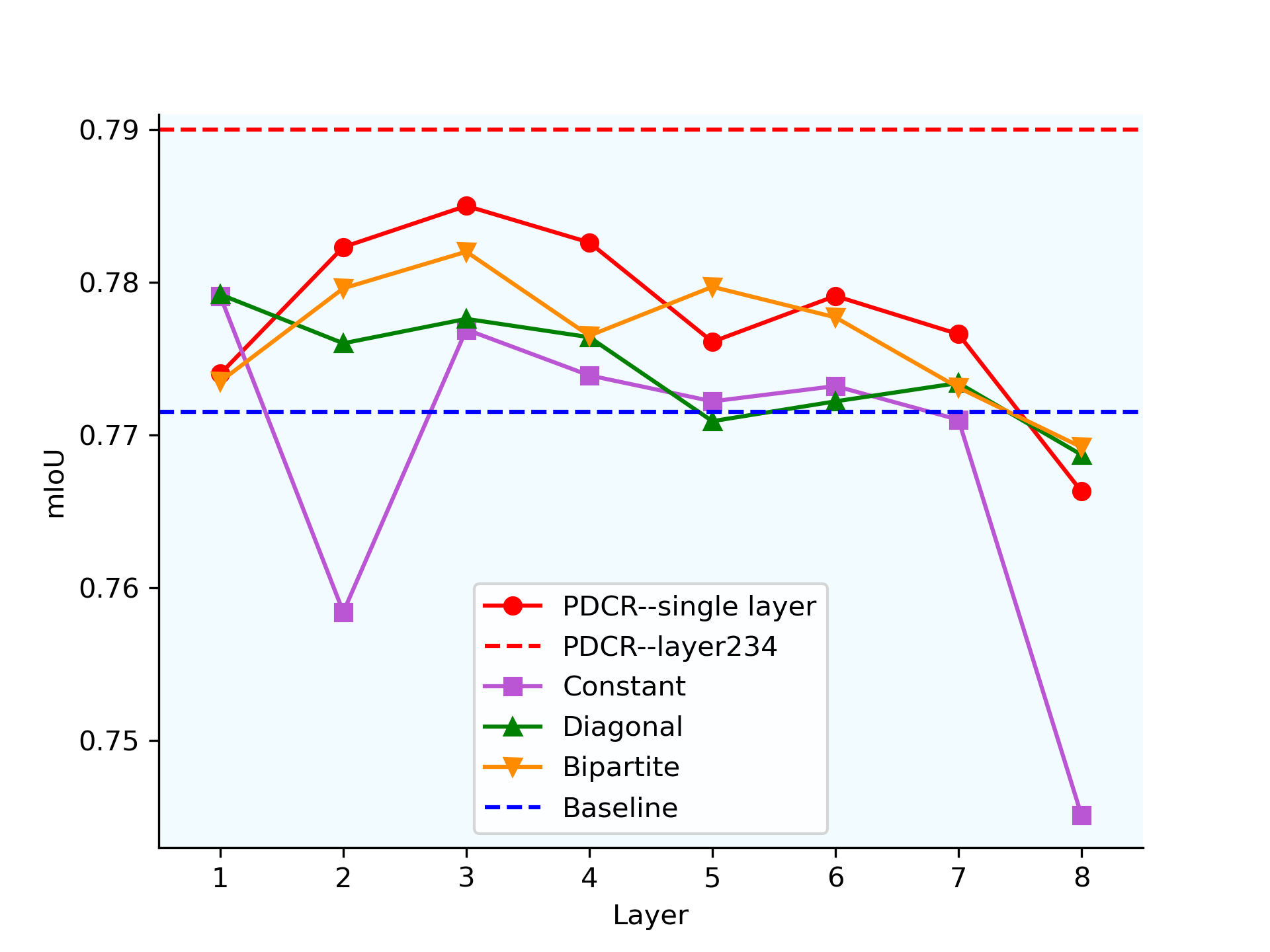}
    \caption{Three variations of affinity score along with PDCR are ablated layer-wisely in the encoder. A clear advantage is gained with PDCR at each layer and the benefit is additive if PDCR is added to layer 2,3,4 simultaneously.}
    \label{img:cl_ab}
\end{figure}

\subsection{Ablation Studies}
To further investigate the benefits gained with the two innovations, the comparison against other designs, and the best configuration, we conduct ablation studies on the ISIC 2017 dataset using DeepLabv3+ with xception backbone. The dataset is chosen for a balance of dataset size and training budget.

\subsubsection{Rule-of-thumb for usage}
We here investigate the best configuration for the two innovations. Specifically, where should the two be plugged in? We take the last layer of each block as well as each singleton layer as candidate sites and ablate the two techniques layer-by-layer, respectively.

\begin{figure}[t]
    \centering
    \includegraphics[width=0.8\linewidth]{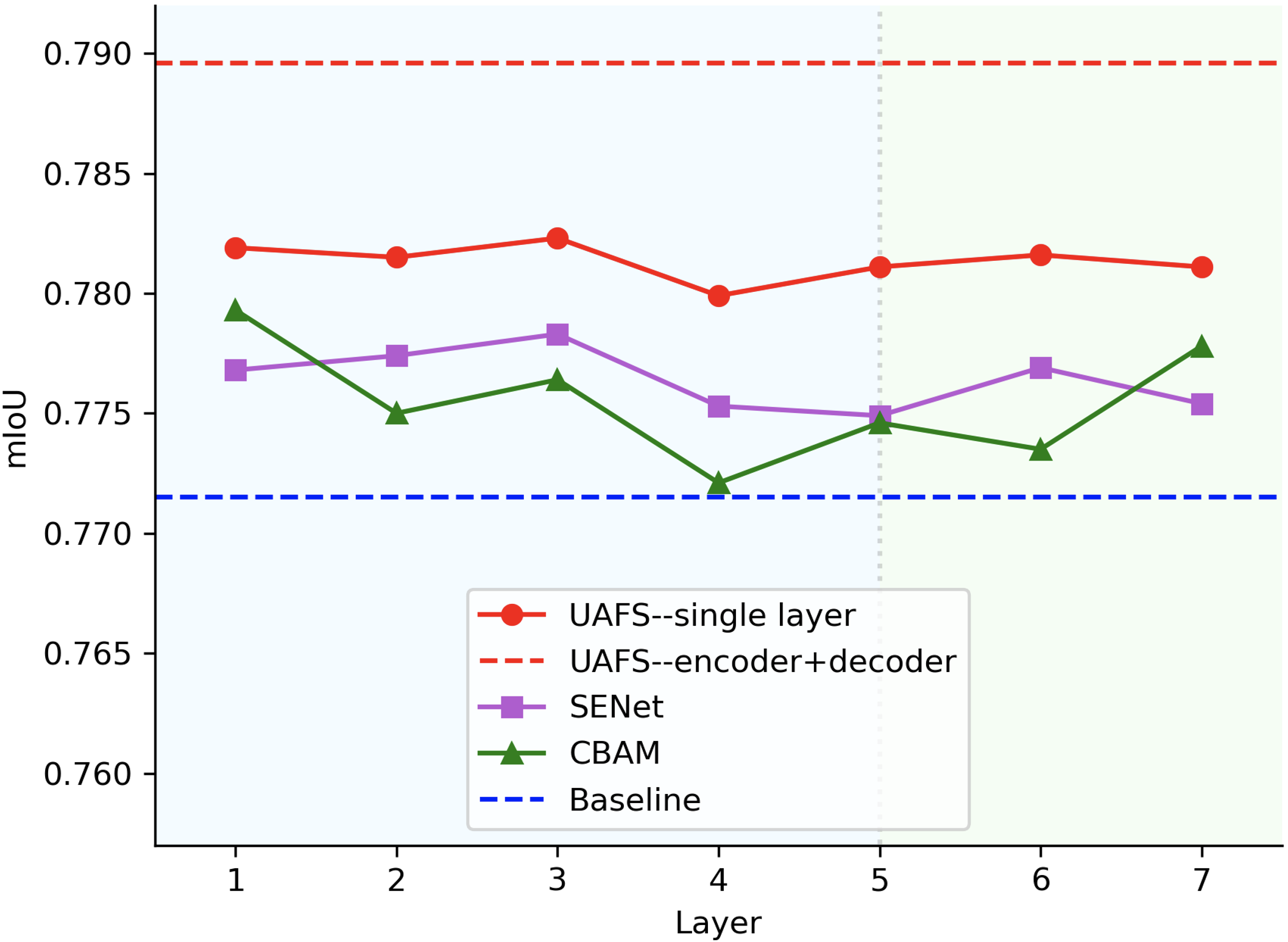}
    \caption{UAFS is comparing with SE-block and CBAM layer-wisely in both encoder (layer 1-5) and decoder (layer 6-7). It's a clear evidence that uncertainty-guided feature selection may function better in medical image domain than attention mechanism. By adding UAFS at all layers further improve the final performance.}
    \label{img:att_ab}
\end{figure}

\noindent
\textbf{Patch-dragsaw.} Patch-dragsaw is limited to encoder layers for the applicability of the receptive field concept, and we leave the extension of this limitation, for example, with a generalization of the receptive field concept for decoder layers, to future works. Fig.~\ref{img:cl_ab} shows the results for this ablation study. It can be clearly seen that patch-dragsaw performs better in the early stage but not the first layer of the encoder. So we add PDCR to layer 2,3,4 simultaneously. The performance is even better, indicating that the benefits are additive. In summary: \textit{use PDCR at layers where the receptive field is neither too large nor too small.}

\noindent
\textbf{Uncertainty-aware feature selection.} As illustrated in Fig.~\ref{img:att_ab}, UAFS is more robust against the plugged-in position and performs roughly at both encoder and decoder. We do not add it at aspp neck since aspp neck is the concatenation of multiple conv layers. Similarly, the benefits are additive and we finally add it to all layers in the encoder and decoder. In a word: \textit{use UAFS at any layer you like in the encoder and decoder.}   

\subsubsection{Similar designs}
\textbf{Variations for affinity score.} Here, we substantiate 3 variations listed as following: (1) \textit{Constant}: we use 0.5 for all entries in the affinity matrix. This essentially downgrades the affinity score to a constant temperature coefficient. (2) \textit{Diagonal}: we set the affinity matrix as a diagonal matrix which indicates a patch is only positive with itself. (3) \textit{Bipartite}: This variation provides a mutual-exclusive division. Bipartite utilizes the label of the center location to represent the whole patch. Then affinity scores for patch-pairs with the same labels are set to 1 and 0 otherwise. According to Fig.~\ref{img:cl_ab}, a clear advantage can be observed, indicating that a continuous affinity score setting may better fit the medical segmentation scenario.

\begin{figure*}[t]
\centering
    \includegraphics[width=0.9\textwidth]{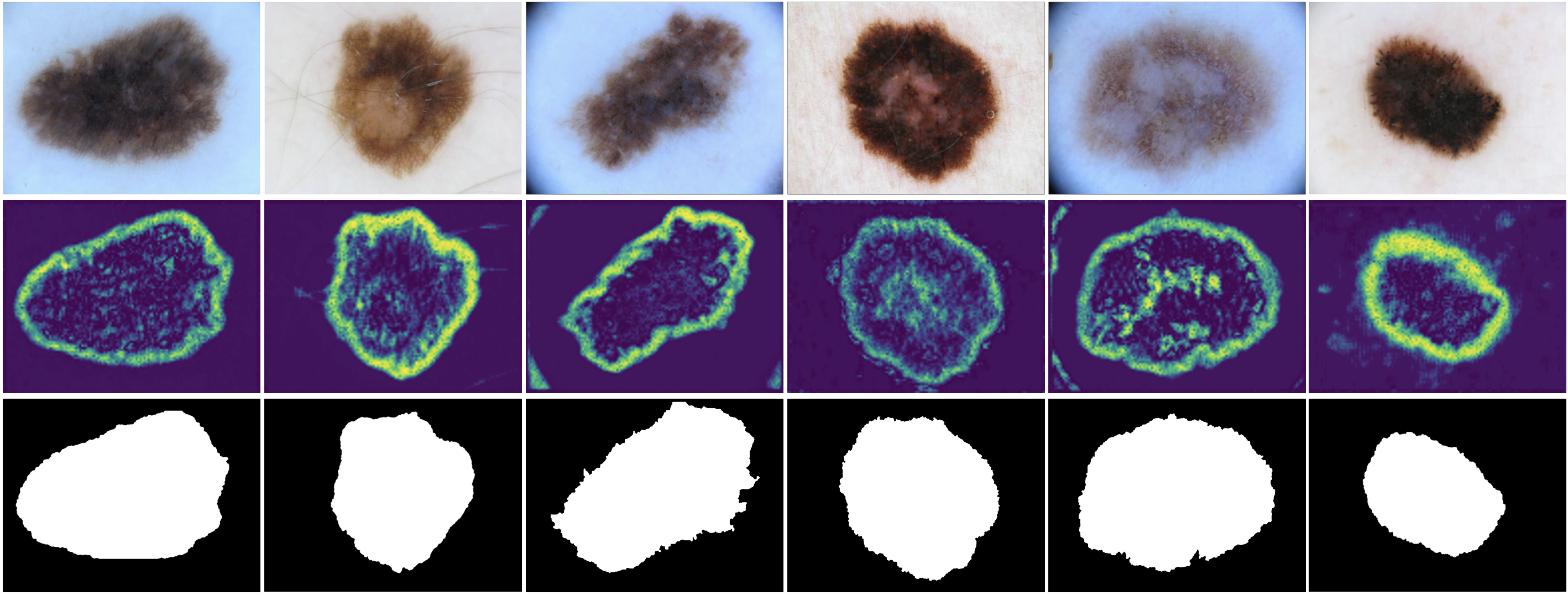}
\caption{Visualization of 6 original images and their corresponding uncertainty maps and ground-truth annotations. This result also satisfy our intuition that obscured boundaries and heterogeneous features receive high uncertainties.}
\label{img:uafs}
\end{figure*}

\noindent
\textbf{UAFS vs Attention.} Uncertainty-aware feature selection can also be viewed as an attention mechanism dedicatedly designed for medical image segmentation scenarios. From this perspective, one may wonder whether other attention mechanisms (e.g. the SE Block) will just suffice. To this end, we replace the proposed structure with two other attention mechanisms, SE block~\cite{SENet} and CBAM~\cite{CBAM}, to see the changes brought to the performance. As seen in Fig.~\ref{img:att_ab} (b), UAFS outperforms two counterparts by a large margin. It is believed that improvements are gained from regions with ambiguous boundaries region as well as heterogeneous textures. Visualization from Fig.~\ref{img:uafs} also substantiate our intuition.

\begin{table}[ht]
\footnotesize
\begin{center}
\begin{tabular}{lccccc}
\toprule
Architecture & +PDCR & +UAFS & JA & DI & AC \\
\midrule
UNet & & & 69.34 & 79.33 & 90.96  \\
UNet &\checkmark& & 70.50 & 80.51 & 92.02 \\
UNet &&\checkmark& 70.28 & 80.41 & 91.94 \\
UNet &\checkmark&\checkmark& \bf 71.35 & \bf 81.52 & \bf 93.01 \\
\midrule
FCN & & &72.31 & 81.20 & 92.22 \\
FCN &\checkmark& & 74.00 & 82.95 & 93.09\\
FCN & &\checkmark& 74.67 & 83.39 & 93.02 \\
FCN &\checkmark&\checkmark& \bf 75.57 & \bf 84.25 & \bf 93.95\\
\midrule
DeepLabv3+ &&& 77.15 & 85.13 & 92.88 \\
DeepLabv3+ &\checkmark& & 79.00& 86.91 & 94.11\\
DeepLabv3+ &&\checkmark& 78.96 & 86.19 & 94.28 \\
DeepLabv3+ &\checkmark&\checkmark& \bf 80.86 & \bf 88.74 & \bf 95.18 \\
\bottomrule
\end{tabular}
\end{center}
\caption{Results of ablation studies for different architectures. A strong and consistent improvement is demonstrated.}
\label{table:arch_ab}
\end{table}

\begin{figure}[t]
\begin{center}
\includegraphics[width=0.9\linewidth]{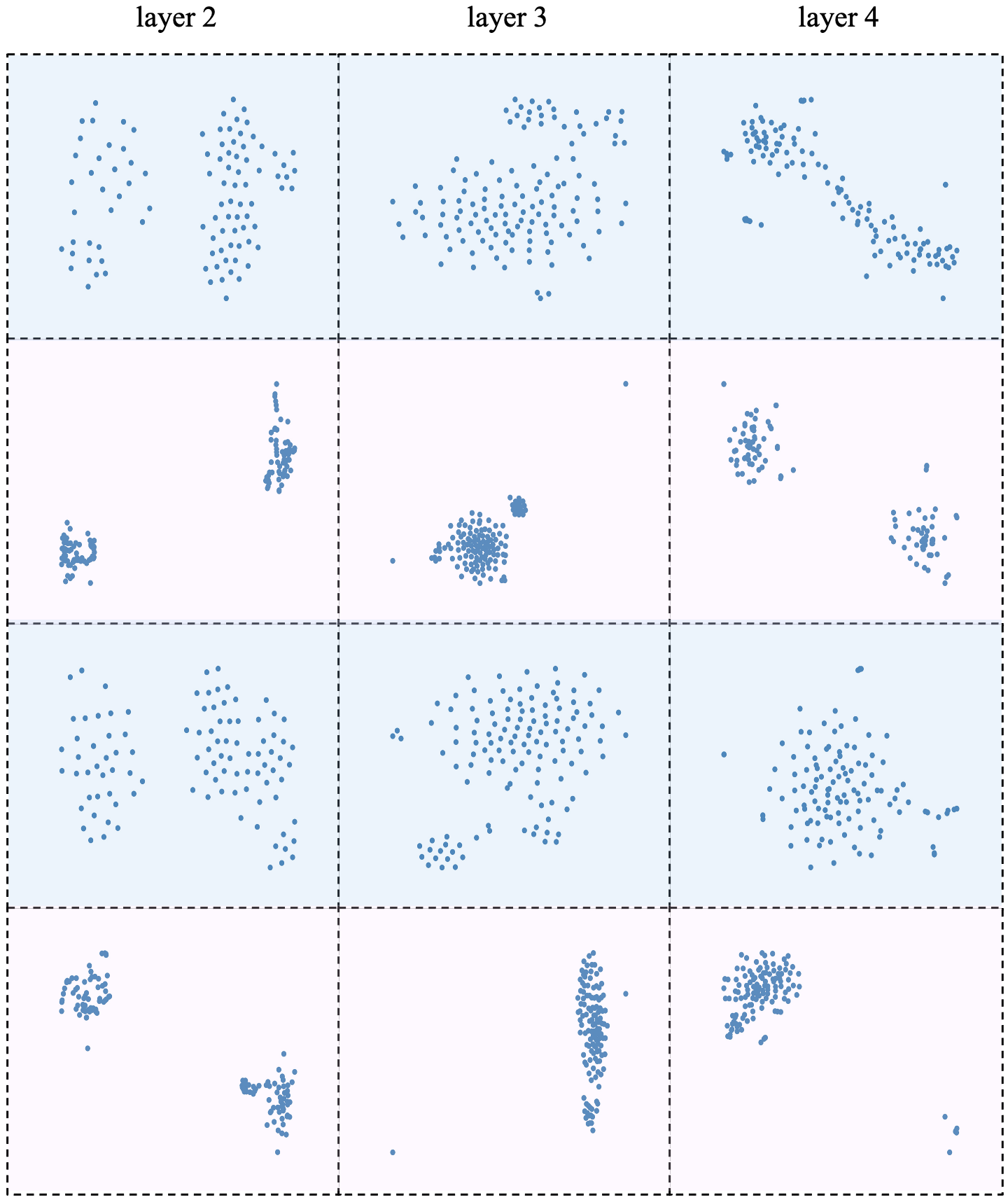}
\end{center}
  \caption{Visualization of 128 sampled hidden vectors from each 2-4 layers using t-SNE~\cite{t-SNE}. Odd-numbered rows are vectors sampled from baseline model and even-numbered rows are counterpart trained with PDCR added at 2,3,4 layers. It's convincing that PDCR helps to increase discrimination ability by encouraging clustering.}
\label{img:pdcr}
\end{figure}

\subsubsection{Plug-and-play}
Patch-dragsaw contrastive regularization and uncertainty-aware feature selection are flexible enough to be plugged into any existing architectures, segmentation frameworks, or training pipelines. In this subsection, we investigate whether the benefits are also transferrable to other architectures. To this end, we add either or both of the two techniques to FCN, UNet, in addition to DeepLabv3+, which are arguably the most common segmentation frameworks among others. Table~\ref{table:arch_ab} shows the results. Two observations are worth noting: Firstly, both patch-dragsaw contrastive regularization and uncertainty-aware feature selection demonstrate consistent improvements on all of the three frameworks, indicating the generalizability; Secondly, the two techniques are complementary in that they are shown to benefit on top of each other.

\subsection{Grab intuitions from visualization}
Qualitative analysis is performed here to facilitate the understanding. For PDCR, we visualize 3 instances by taking 128 sampled hidden vectors from each 2-4 layers of the network with or without the technique, and then we use t-SNE~\cite{t-SNE} to visualize the distribution of sampled vectors from each layer. As shown in Fig.~\ref{img:pdcr}, odd-numbered rows represent the baseline network, and even-numbered rows are network trained with PDCR. The comparison clearly shows that PDCR is effective in discriminating different hidden vectors through clustering and it is owing to, as we think, the power of continuously controlled contrastive regularization. 
For UAFS, we illustrate uncertainty maps in Fig.~\ref{img:uafs}. Regions with ambiguous boundaries or heterogeneous textures are clearly highlighted in uncertainty maps, meaning that regions are decided by the network to be uncertain and thus receive lower weights. Since uncertainty maps in the model do not receive any direct supervision and are mainly driven by the overall objectives, the phenomenon demonstrates that the design of the structure is just suitable for modeling the uncertainty.
%------------------------------------------------------------------------
\section{Conclusion}
In this work, we propose to manipulate hidden embeddings at each layer by imposing contrastive constraints and selective weights. Patch-dragsaw contrastive regularization explores relations between vectors and their collateral patches. Uncertainty-aware feature selection performs vector-wise selection according to the uncertainty modeled by network itself. Promising results have been shown on both supervised learning and limited-data scenario. For future work, we plan to further explore the functionality of the proposed two techniques in un-/semi-supervised learning. In addition, it is more convenient if the limitation caused by receptive field can be decoupled without sacrificing overall performance. 

{\small
\bibliographystyle{ieee_fullname}
\bibliography{egbib}
}
\end{document}